\documentclass[a4paper]{llncs}
\usepackage[utf8]{inputenc}

\usepackage[dvipsnames]{xcolor}
\usepackage{graphicx}
\usepackage[caption=false]{subfig}
\usepackage{mathtools}
\usepackage{listings}
\usepackage{todonotes}
\usepackage{comment}
\usepackage{hyperref}
\usepackage{multirow}
\usepackage[normalem]{ulem}
\usepackage{subfiles}
\usepackage{csquotes}
\usepackage[noend]{algpseudocode}
\usepackage{algorithm}

\begin{document}

\title{LazyBum: Decision tree learning using lazy propositionalization}

\author{
	Jonas Schouterden\orcidID{0000-0003-3071-7259} 
	\and Jesse Davis\orcidID{0000-0002-3748-9263}
	\and Hendrik Blockeel\orcidID{0000-0003-0378-3699} 
}

\institute{KU Leuven, Department of Computer Science\\
	Celestijnenlaan 200A box 2402, 3001 Leuven, Belgium\\
	\email{\{jonas.schouterden,jesse.davis,hendrik.blockeel\}@cs.kuleuven.be}
}

\maketitle

 \begin{abstract}
Propositionalization is the process of summarizing relational data into a tabular (attribute-value) format.  The resulting table can next be used by any propositional learner. This approach makes it possible to apply a wide variety of learning methods to relational data.  However, the transformation from relational to propositional format is generally not lossless: different relational structures may be mapped onto the same feature vector.  At the same time, features may be introduced that are not needed for the learning task at hand.  In general, it is hard to define a feature space that contains all and only those features that are needed for the learning task.
This paper presents LazyBum, a system that can be considered a lazy version of the recently proposed OneBM method for propositionalization.  LazyBum interleaves OneBM's feature construction method with a decision tree learner. This learner both uses and guides the propositionalization process. It indicates when and where to look for new features.  This approach is similar to what has elsewhere been called dynamic propositionalization. In an experimental comparison with the original OneBM and with two other recently proposed propositionalization methods (nFOIL and MODL, which respectively perform dynamic and static propositionalization), LazyBum achieves a comparable accuracy with a lower execution time on most of the datasets.
\end{abstract}

\section{Introduction}
There is a  renewed interest in analyzing data stored in relational databases.  In 2017, Tan et al.\ proposed the “One Button Machine”  (OneBM)~\cite{Lam2017}, which automatically constructs features from a relational database.  In ILP terms, one would say that OneBM performs propositionalization~\cite{Lachiche2017}.  It summarizes a relational database into a single table by defining features that are derived from the database by joining multiple tables.  It handles one-to-many and many-to-many relationships by using specific aggregation functions that aggregate the information in a set of multiple related tuples into a single tuple.

An obvious disadvantage of propositionalization is that there is usually a loss of information: the resulting table provides a summary of the original database, from which that database cannot uniquely be reconstructed.  Defining more features means that less information is lost.

Viewed from an ILP perspective, propositionalization is equivalent to defining a (usually relatively small) set of clauses, and associating with each clause one particular feature.  A typical ILP system searches a space that is much larger than the number of features typically constructed by propositionalization approaches.

In this paper, we propose a variant of OneBM that performs dynamic, or ``lazy'', propositionalization. It considers the same types of features as OneBM, but constructs these features in a lazy manner that is guided by the learner. We begin by only considering that are based on the target table. But when another table's relevance to the learner becomes more evident, it expands its feature space to consider features based on information contained in that table. 

The gradual expansion of the feature table is somewhat similar to how ILP systems gradually construct longer clauses by first constructing shorter ones and considering only the promising ones for extension.  An important difference, however, is that ILP systems, when evaluating a clause, typically re-evaluate the whole clause, which includes re-discovering answer substitutions for the subclause that has already been evaluated earlier. The lazy propositionalization methods proposed in this paper caches these instantiations.

The hypothesis underlying this paper is that a method like OneBM can be made more efficient in both memory and time by implementing a lazy version of its feature construction, without a loss of accuracy.  At the same time, one might hope that it is faster than ILP systems that use the same implicit search space.

The remainder of this paper is structured as follows. Section~\ref{sec:onebm} briefly presents OneBM. Section~\ref{sec:lazybum} introduces our new algorithm, including two available strategies for defining new features, and discusses related work. Section~\ref{sec:evaluation} experimentally compares this algorithm to other propositionalization approaches, in terms of predictive and run-time performance, and Section~\ref{sec:conclusion} presents conclusions.

\section{OneBM}
\label{sec:onebm}

\begin{figure}
	\centering
	\includegraphics[width=0.9\textwidth]{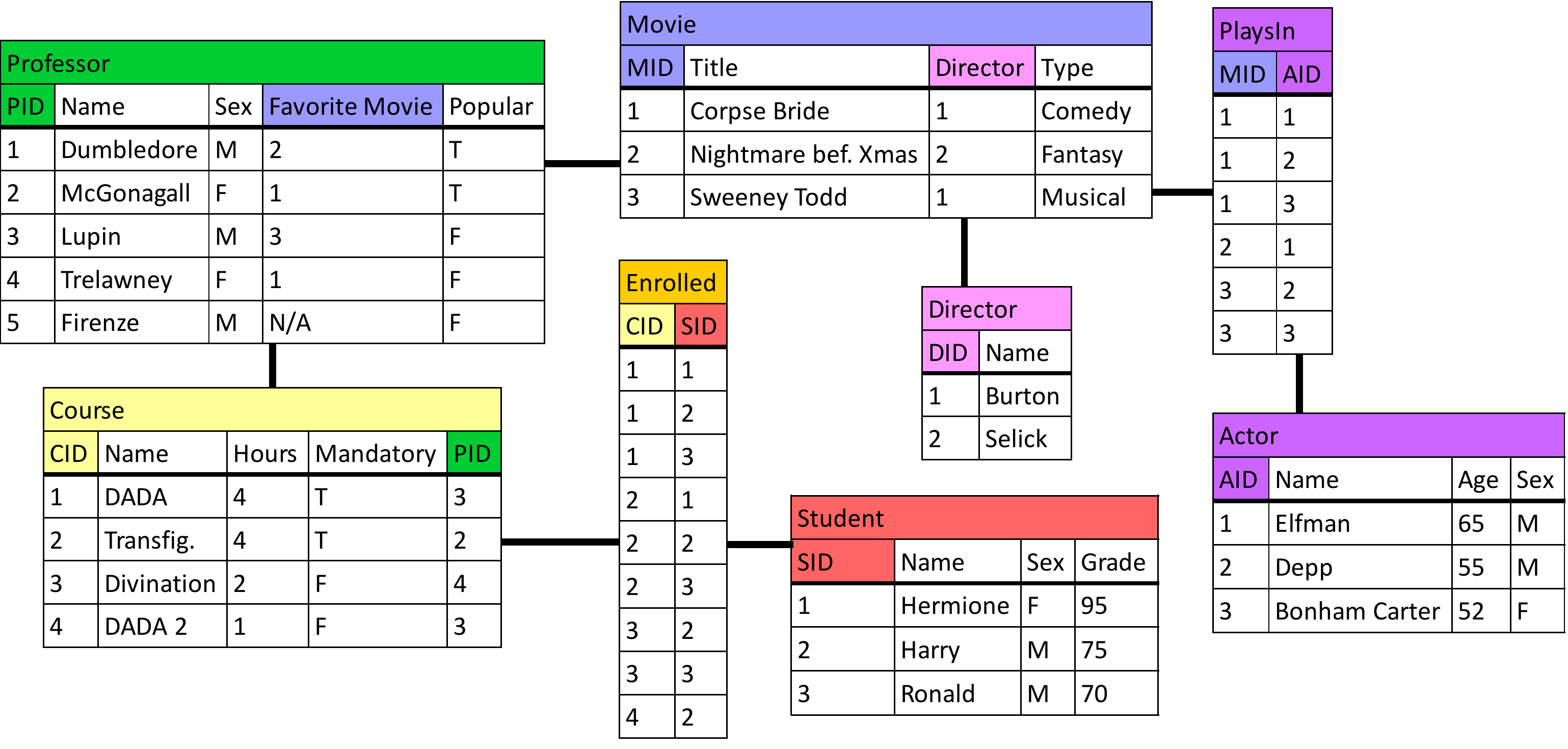}
	\caption{An example database representing teachers and students in a fictional school. \emph{Professor} is the target table and the target attribute is \emph{popular}.}
	\label{fig:running_example_lupin}
\end{figure}

The ``One Button Machine'' or OneBM~\cite{Lam2017} is a relational learning system that works on data stored in a relational database.  It takes as input a set of tables, connected with each other through foreign keys. A single attribute is selected to serve as the target attribute, and the table containing this attribute is called the target table.
OneBM produces as output a modified target table that contains newly constructed features which summarize the other tables.
Figure~\ref{fig:running_example_lupin} shows an example of what the input may look like.

The OneBM paper defines a ``joining path'' as a sequence of tables $T_0 \xrightarrow{c_1} T_1 \xrightarrow{c_2} T_2 \xrightarrow{c_3} \cdots T_k \mapsto {A}$ where $T_0$ is the target table, $c_i$ is the condition on which $T_{i-1}$ and $T_i$ are (equi-)joined, and $A$ is an attribute of the last table in the sequence.  Note that this definition considers the projection onto one single attribute at the end as part of the ``joining path.'' In this paper, we will use the term ``join path'', or J-path, for the joining path without the final projection, and ``join-project path'', or JP-path, for the joining path as originally defined.
Given a $T_0$-tuple $t$, we will write $t.P$ for the set of $T_k$-tuples associated with it through the join path $P$, and $t.P.A$ for the multiset of $A$-values in $t.P$. 

\begin{figure}
	\centering
	\includegraphics[width=0.5\textwidth]{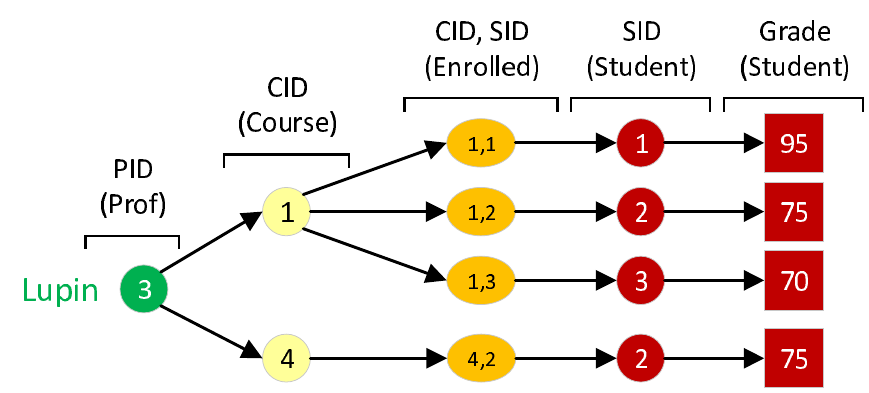}
	\caption{The multiset of grades for students taking one of professor Lupin's courses. OneBM transforms this multiset to multiple features for Lupin. Example transformations include the average and standard deviation.}
	\label{fig:GradesStudentsCoursesLupin2}
\end{figure}

Figure~\ref{fig:GradesStudentsCoursesLupin2} shows, for the database shown in Figure~\ref{fig:running_example_lupin} and the JP-path  
\(\mathit{Professor}\allowbreak \xrightarrow{\mathit{PID}} \mathit{Course} \xrightarrow{\mathit{CID}} \mathit{Enrolled} \xrightarrow{\mathit{SID}} \mathit{Student} \mapsto \mathit{Grade}\), the multiset of grades associated with Prof. Lupin (we use $\mathit{PID}$ as shorthand notation for $\mathit{Professor}.\mathit{PID} = \mathit{Course}.\mathit{PID}$ here, and similar for $\mathit{CID}$ and $\mathit{SID}$).

If all the joins in $P$ are one-to-one or many-to-one, then $t.P.A$ is guaranteed to be a singleton; otherwise, it is not.  In the first case, we call $P$ determinate, and in the second case we call it non-determinate.

OneBM derives features from JP-paths as follows.  A determinate path defines one feature, whose value (for a given tuple $t$) is the single element of $t.P.A$.  
A non-determinate path defines a fixed-sized feature vector whose components are defined by predefined aggregation functions applied to $t.P.A$.  Which aggregation functions are used depends on the type of $A$.  If $A$ is numerical, the feature vector contains the mean, variance, min, max, sum and count of the numbers in the multiset.  If $A$ is categorical, the feature vector contains the cardinality of the multiset and the corresponding set (in SQL terms, the count and count distinct functions). OneBM defines other aggregation functions for values that are texts, timestamps, etc.  

The features defined by a JP-path can be collected using a single SQL query.  For example, the multiset from Figure~\ref{fig:GradesStudentsCoursesLupin2} and its corresponding features can be computed using the following SQL query:

\begin{verbatim}
SELECT count(grade), sum(grade), average(grade), 
       variance(grade), min(grade), max(grade)
FROM Professor 
     JOIN Course ON Professor.PID = Course.PID
     JOIN Enrolled ON Course.CID = Enrolled.CID
     JOIN Student ON Student.SID = Enrolled.SID
GROUP BY Professor.PID
\end{verbatim}

OneBM defines the depth of a table $T_i$, $d(T_i)$, as the length of the shortest join path between $T_0$ and $T_i$.  It has two options for generating join paths: in ``forward-only'' mode, consecutive tables must be increasingly farther from the target table (that is, $i<j \Rightarrow d(T_i) < d(T_j)$), whereas in ``full'' mode this restriction is dropped, allowing for what the authors of OneBM call ``backward traversal''. 

Given a database, OneBM constructs a table that contains for each join path all the defined features. As the number of join paths can grow exponentially with their length, OneBM has a MaxDepth parameter that limits this length.

OneBM uses a relatively restrictive bias.  For instance, it does not mix selections into the join path, as, e.g., Van Assche et al.\ do \cite{VanAssche2006}.  Doing so would result in an exponential blowup of the already large feature table.

The lazy version of OneBM that we propose does not attempt to lift this restriction or address any other limitations; it constructs features lazily, but in all other respects is meant to behave as much as possible like OneBM.  It is currently limited to numerical and categorical values, and only implements the forward-only approach.

\section{LazyBum}
\label{sec:lazybum}

Our lazy version of OneBM is called Lazy Button Machine, or LazyBum.
The main motivation for developing LazyBum is that, by constructing all features in advance, OneBM may invest much work into computing features that afterwards will not be used by the learner.  LazyBum takes a more cautious approach: it first computes features based on short join paths, and only extends join paths when (1) the simpler features turn out to be insufficient, and (2) there is reason to believe the extension will help.  In principle, LazyBum can construct every feature that OneBM can construct, as it explores the same feature space.  In practice, it avoids constructing the large majority of them.

As the lazy feature construction somehow needs to be informed about which tables are useful additions to the current path, it is natural to integrate a learning system into the feature construction process.  LazyBum is based on relational decision tree learning (such as Tilde~\cite{Blockeel:1998:TIF:1643275.1643308} or Relational Probability Trees~\cite{Neville2003}).

\subsection{The LazyBum Algorithm}

 LazyBum learns a decision tree in a top-down fashion. When deciding on the test to include in a node, it evaluates possible tests using a ``local data table'' (LDT), which just like in traditional tree learning contains all instances sorted into this node, with one row per instance. Unlike traditional tree learning, the LDT does not have a fixed schema (set of attributes) as the learner may extend the schema with new features as needed. Each LDT is associated with a decision tree node and contains all the features constructed along the path from the root to that node. We now explain the learning process in detail.

 At the root node of the tree, the LDT is the target table extended with all features derived from join paths of length 1 (i.e., the tables directly connected to the target table). LazyBum uses information gain to select the most informative feature in the LDT to split on. If a good enough split is found, it splits the LDT into two subsets of rows based on this: one for each child. If no good split can be found, LazyBum tries to extend the LDT by introducing new features.
 These new features are defined by extending some of the join paths used to build the current LDT. Information from the current decision tree branch can guide the selection of which join paths to extend, which is discussed in Subsection~\ref{subsec:LDT-extension-strategies}.
 LazyBum finds the best split based on the new features, and splits the extended LDT into two subsets. If none of the new features is good enough, the node is turned into a leaf.  This procedure is recursively repeated for all subsets created.  Algorithm~\ref{alg:main-LazyBum} summarizes the entire procedure. 
 
\begin{algorithm}[ht]
	\caption{Main LazyBum algorithm.}
	\label{alg:main-LazyBum}
	\begin{algorithmic}[1] 
		\Require
			\Statex $\mathit{MaxDepth}$, max tree depth, 
			\Statex $\mathit{MinInst}$, minimum nb of instances in a leaf
			\Statex $\mathit{MinIG}$, minimum information gain threshold
		
		\Procedure{grow\_tree}{node $N$, table $\mathit{LDT}$} 

		\If{ depth($N$) $= MaxDepth$ \textbf{or} $\#$rows($LDT$)  $< \mathit{MinInst}$}
			\State Make $N$ a leaf node
		\Else
			\State Find the test $\tau$ with highest information gain to split $\mathit{LDT}$ on
			\If{IG($\tau$) $> \mathit{MinIG}$}	
				\State Split $\mathit{LDT}$ into tables $LDT_L$, $LDT_R$
				\State Turn $N$ into an inner node with children $N_L, N_R$
				\State Call grow\_tree($N_L, LDT_L$) and grow\_tree($N_R, LDT_R$)
			\Else
				\If{table $\mathit{LDT}$ can be extended}
					\State $LDT^{ext}\ \gets \text{extend\_data\_table}(\mathit{LDT}, N)$
					\State Find the test $\tau$ with highest information gain to split $LDT^{ext}$ on
					\If{IG($\tau$) $> \mathit{MinIG}$}
						\State Split $LDT^{ext}$ in tables $LDT^{ext}_L$, $LDT^{ext}_R$
						\State Make $N$ an inner node with children $N_L, N_R$
						\State Call grow\_tree($N_L, LDT^{ext}_L$) and grow\_tree($N_R, LDT^{ext}_R$)
					\Else 
						\State Make $N$ a leaf node
					\EndIf
				\Else \Comment{$\mathit{LDT}$ cannot be extended}
					\State Make $N$ a leaf node
				\EndIf
			\EndIf
		\EndIf
		\EndProcedure
	\end{algorithmic}
\end{algorithm} 

 The way LDTs are extended is somewhat similar to the way in which the relational decision tree learner Tilde extends its clausal queries.  In Tilde, a query $Q$ is associated with the current node ($Q$ contains all the tests from the root to this node), and this query is extended with one or more literals, chosen among many candidates. After the most informative extension $e$ is found, the set of instances satisfying $Q$ (i.e., all instances at this node) is partitioned into a subset of instances that satisfy $Q \land e$, and a subset of instances that do not. Important differences between Tilde and LazyBum are:
 \begin{itemize}
 	\item In Tilde, for each candidate extension $e$, the query $Q \land e$ is evaluated. This means the sub-query $Q$ is computed many times. The ``query pack'' implementation of Tilde \cite{Blockeel2002EfficiencyThroughQueryPacks} avoids this to some extent: within a single node, the search for all answer substitutions for query $Q$ is done only once, not once for each extension.
 	
	While query packs avoid rerunning $Q$ multiple times inside one node, $Q$ must still be rerun in that node's children.
	LazyBum differs in this respect.
	LazyBum caches the join paths corresponding to a node's LDT.  A join path is cached using the primary keys identifiers of the tables on its path, for the instances in the LDT. When extending a node's LDT, it reuses the join paths of its ancestor nodes to avoid recomputing these joins. Only the joins with the extension tables need to be calculated. This is similar to caching all answer substitutions of $Q \land e$ for all possible extensions $e$, and reusing the cached results in all child nodes. 
	
 	 	 	
    In addition, when extending a LDT, LazyBum derives all features for each join path extension and adds them to the LDT. Therefore, if a feature is relevant but not chosen immediately because a better feature exists, this feature will appear in the LDT for  all descendant nodes and hence can be used as split criteria in one of these nodes (without having to be recomputed). In comparison, Tilde with query packs does not cache the `features' it does not split on for use in child nodes. 
 
 	\item Tilde uses a more flexible language bias, largely specified by the user, whereas LazyBum uses a predefined and more restrictive bias. LazyBum's bias is intended to be restrictive enough to make the storage of the LDT feasible.
 \end{itemize}
  
\subsection{LDT Extension Strategies}
\label{subsec:LDT-extension-strategies}
 
We call two tables {\bf neighbors} if they are connected by a foreign key relationship.

LazyBum defines two different strategies for extending an LDT. 
Let $P = T_0 \rightarrow T_1 \rightarrow \cdots \rightarrow T_k$ be a J-path used to construct the LDT.
A table is called a candidate for extension of $P$ if (a) it does not occur in $P$ and (b) it neighbors on $T_k$.
In the {\bf unrestricted} strategy, every path used to construct the current LDT gets extended with each candidate for extension.
In the {\bf restricted} strategy, only those paths get extended from which at least one feature actually occurs in an ancestor node of the decision tree node currently being considered. Hence, the decision tree guides the selections of which join paths should be extended.
The difference between the two strategies is that for the unrestricted strategy, it suffices that the features defined by the join path have been introduced in the LDT, while for the restricted strategy they must also have been used at least once. The motivation for the latter condition is that LazyBum should preferentially introduce relevant features, and the underlying assumption is that tables are more likely to be relevant if their neighbors are. 

The LDT is then extended by considering for each extended J-path all JP-paths (that is, considering each attribute of the newly added tables), computing the features defined by these JP-paths, and adding these features to the LDT. Table~\ref{tab:feature-transformations} lists the aggregation functions that are currently used by LazyBum.
Most of them speak for themselves.  The ``contains'' aggregation function introduces for each possible value of a categorical domain a Boolean feature that is true if and only if the value occurs in the multiset. To avoid problems with ``categorical'' variables that have a very large domain (e.g., because they are in fact strings), these features are only introduced for variables whose domain size is below both an absolute threshold \(\mathit{DomSize}_{abs}\) and a relative threshold \(\mathit{DomSize}_{rel}\) (relative to the number of rows in the table).

\begin{figure}
	\centering
	\includegraphics[width=0.8\textwidth]{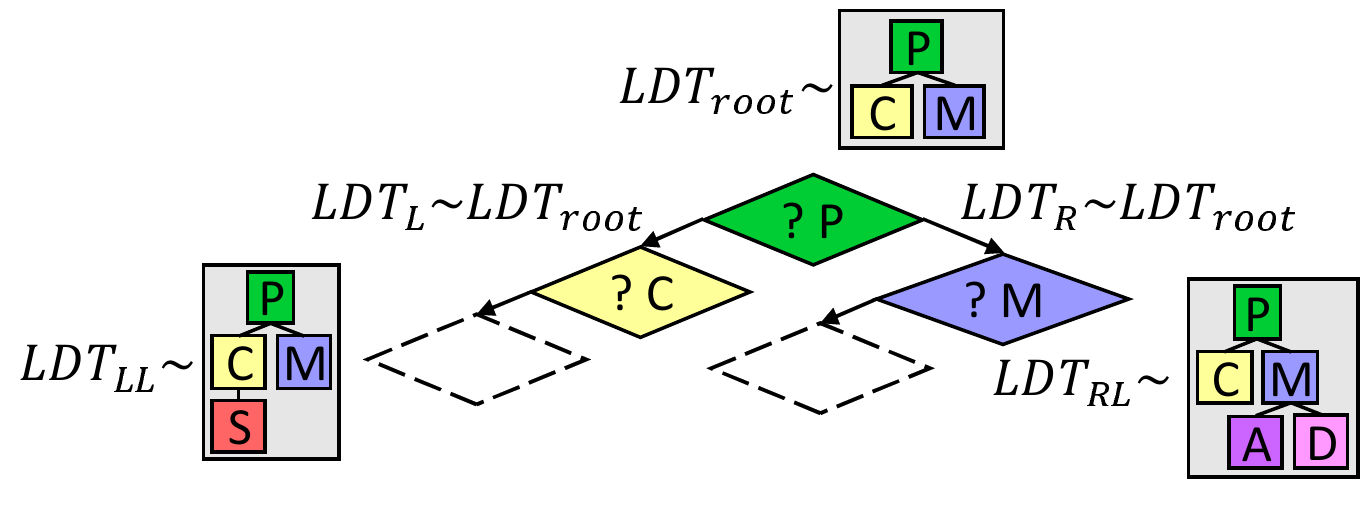}
	\caption{Example of a LDT schema extension in the restricted strategy based on the splits chosen in the decision tree nodes, using the example database from Figure~\ref{fig:running_example_lupin}. The partial decision tree shows the root node, its left child and its right child splitting on features from \(\mathit{Professor}\), \(\mathit{Professor} \xrightarrow{\mathit{PID}} \mathit{Course}\) and \(\mathit{Professor} \xrightarrow{\mathit{MID}} \mathit{Movie}\), respectively. The schemas of the LDTs of the root and its two children are the same. For both left children of the roots two children, no good split is found and their LDTs are extended. }
	\label{fig:running_example_LDT_extensions}
\end{figure}

Figure~\ref{fig:running_example_LDT_extensions} illustrates how the schemas of LDTs are extended in the restricted strategy, guided by their corresponding decision tree branches.

\begin{table}[ht]
\begin{tabular}{|l|l|l|}
	\hline
	Join result                   & Data type   & aggregation function                                          \\ \hline
	\multirow{2}{*}{single value} & numerical   & \multirow{2}{*}{identity function}                      \\ \cline{2-2}
	& categorical &                                                         \\ \hline
	\multirow{2}{*}{multi-set}    & numerical   & avg, standard deviation, variance, max, min, sum, count \\ \cline{2-3} 
	& categorical & count, distinct count, contains                   \\ \hline
\end{tabular}
\caption{The functions supported by LazyBum to aggregate multisets.}
\label{tab:feature-transformations}
\end{table}

\subsection{Special cases}

\paragraph{Associative tables}

LazyBum normally extends join paths with a single table, but there is one exception to this rule. Many-to-many relationships are usually implemented with an intermediate associative table that does nothing else than connect two other tables in an m-to-n manner (for an example, see the table Enrolled in Figure~\ref{fig:running_example_lupin}).  The intermediate table has no attributes of its own, and is more of an implementation artifact than a conceptual table. For this reason, when an intermediate table is selected, the tables it connects to is immediately added as well.  This is somewhat similar to lookahead in ILP.

\paragraph{Empty multisets} For some tuples $t$ and join paths $P$, $t.P.A$ may be empty. Not all aggregation functions are defined on the empty set.  To deal with this, LazyBum uses the following strategy when evaluating features to split on. For any test, the instance set is split into three subsets Pass, Fail, and Undefined, which respectively contain the instances that pass the test, fail the test, or are untestable because the feature is undefined. This ternary partition is transformed into a binary partition by merging Undefined with either Pass or Fail, depending on which of these two yields the highest scoring test. If that test is eventually chosen, the node stores which option was chosen, so that it can correctly handle instances with undefined values at prediction time.  Apart from this, LazyBum also introduces a Boolean feature that indicate whether a multiset is empty or not, as this by itself may be relevant information.

\paragraph{Missing values} Missing values may occur in the input data.  Missing values are quite different from undefined values, and must be treated differently.
When the original database has missing values, these are included as separate elements in $t.P.A$.  Except for count and count distinct, all aggregation functions are computed on the sub-multiset of the multiset that excludes missing values.  When that sub-multiset is empty, while the multiset itself is not, a default value is included as the feature's value.

\subsection{Comparison with related work}
\label{sec:related-work}	

Many propositionalization approaches have been proposed in the past \cite{Krogel2005,Lachiche2017,Boulle2018}.  To better position LazyBum with respect to the state of the art, we categorize them along four dimensions.

\paragraph{ILP vs Databases}
A first dimension is the perspective that they take. Some approaches take an ILP-based first-order logic perspective, other approaches take a relational database perspective~\cite{Krogel2003}. Although the logical and database representations are strongly related, logic and database query engines are typically optimized for different types of queries (essentially, determining the existence of at least one answer substitution, versus computing the set of all answer substitutions). LazyBum is set in the database setting.  The motivation for this is that the features it computes are indeed based on entire sets of answer substitutions.

\paragraph{Types of features}
We distinguish among three types of features.  The first type is existential features, which simply check the existence of an answer substitution of a particular type. These features are typically constructed by propositionalization approaches that are closely related to ILP learners. Examples of such systems are LINUS~\cite{Lavrac1991}, DINUS~\cite{Lavrac:1993:ILP:562956}, SINUS~\cite{Krogel2003}, RSD~\cite{Zelezny2006}, RelF~\cite{Kuzelka2011} and nFOIL~\cite{Landwehr2005}. They often focus on categorical attributes, with numerical attributes getting discretized. 

The second type of features is based on simple aggregation functions, which summarize information in neighboring tables~\cite{Perlich:2003:AFI:956750.956772}. Most initial relational-database oriented propositionalization approaches focus on this type of feature. Examples of such systems are POLKA~\cite{Knobbe2007}, RELAGGS~\cite{10.1007/3-540-44797-0_12}, Deep Feature Synthesis~\cite{Kanter2015} and OneBM~\cite{Lam2017}. 

The third type of features consists of complex aggregates~\cite{Vens2004,Vens2008}. A complex aggregate combines an aggregation function with a selection condition. The ILP learners Tilde and FORF included in the ACE system~\cite{VanAssche2006} allow for complex aggregates to be used. A recent propositionalization approach that considers complex aggregates is MODL~\cite{Boulle2014,Boulle2018}, which is included in the Khiops data mining tool. MODL was designed to deal with a possibly infinite feature space. The approach it takes is two-fold. First, it postulates a hierarchical prior distribution over all possible constructed features. This prior distribution penalizes complex features. It takes into account the recursive use of feature construction rules\footnote{A constructed feature can be used as an argument for another construction rule.}, being uniform at each recursion level. Second, it samples this distribution to construct features.

LazyBum does not construct complex aggregates, but focuses on simple aggregates as used in OneBM. However, LazyBum does build some features using the existential quantifier. 

\paragraph{Indirectly linked complementary tables} This dimension is specific to database-oriented propositionalization approaches and concerns how they handle complementary tables that are not directly connected to the target table. Like OneBM, LazyBum  joins tables over a path through the database, aggregating information for each instance using a single aggregation function. In contrast, POLKA and Deep Feature Synthesis use aggregation functions recursively, aggregating in between joins. RELAGGS uses a form of identifier propagation similar to CrossMine~\cite{DBLP:conf/icde/YinHYY04} to directly relate all complementary tables to the target table.

\paragraph{Static vs. dynamic propositionalization} Static propositionalization approaches perform the following two-step process: (1) convert the relational database to  a data table, and (2) apply any propositional learner to the data table. In contrast, dynamic approaches~\cite{Davis2005,Davis2007a,Landwehr2005} interleave feature construction and model learning. LazyBum is a dynamic version of OneBM, constructing a data table gradually, as needed. LazyBum differs from existing dynamic propositionalization systems like  SAYU~\cite{Davis2005,Davis2007a,davis.icml07} and nFOIL~\cite{Landwehr2005} in three important ways. First, LazyBum takes a database perspective, whereas  SAYU and nFOIL come from an ILP-perspective. Second, LazyBum considers a much wider array of aggregations whereas prior approaches focus on existence~\cite{Davis2005,Landwehr2005} or possible simple counts~\cite{Davis2007a}. Finally, LazyBum guides the propositionalization by learning a decision tree, while nFOIL and SAYU use Bayesian network classifiers.

\section{Evaluation}
\label{sec:evaluation}

The goal of the empirical evaluation is to compare LazyBum's predictive and run-time performance to that of other propositionalization approaches.

\subsection{Methodology}
\begin{table}[htbp]
	\centering
	\caption{The datasets used in the experiments.}
	\label{tab:data-sets-overview}
	\begin{tabular}{|l|l|l|l|l|}
		\hline
		& Hepatitis & UW-CSE 	& University 	& IMDb \\ \hline
		\# examples			& 500 		& 278 		& 38 			& 12000 \\ \hline
		\# classes			& 2 		& 4 		& 3 			& 3 \\ \hline
		\# rows (in total)	& 12927 	& 712 		& 145 			& 442698 \\ \hline
		\# tables 			& 7 		& 5			& 5 			& 8 \\ \hline
		target table 		& dispat 	& person 	& student		& movies \\ \hline
		target variable 	& type		& inphase   & intelligence 	& rating \\ \hline
	\end{tabular}
\end{table}

The following datasets were used in the evaluation, which were collected from the CTU Prague Relational Dataset Repository \cite{DBLP:journals/corr/MotlS15}:
\begin{itemize}
	\item The Hepatitis dataset describes patients with hepatitis B and C. The goal is to predict the type of hepatitis.
	\item The UW-CSE dataset contains information about the University of Washington's computer science department. The goal is to predict the phase a person is in.
	\item The University dataset is a small dataset containing information about students. The classification task is to predict the intelligence of a student. 
	\item The IMDb (Internet Movie Database) dataset contains information relating movies, directors and actors. A possible regression task is to predict the rating of a movie, which is a real number between 0 and 10. We turned this into a classification problem by divided the examples into 3 groups: those with a rating lower than 3.3 (bad movies), those with a rating between 3.3 and 6.6 (average movies) and those with a rating higher than 6.6. The original dataset contained 67245 instances, with 5219 bad, 39599 average and 22427 good movies. From this dataset, we sampled 12000 examples, with 4000 examples of each class.
	
\end{itemize}
The datasets vary in size and number of instances (Table~\ref{tab:data-sets-overview}). For each dataset, we removed the feature columns from the main table, leaving only the primary key and the target attribute (and foreign keys). This ensures that the systems must use information from the secondary tables to result in a model that performs better than a majority class predictor. 

We compare two versions of LazyBum (using respectively the restricted and unrestricted strategy) to the following alternative approaches:
\begin{itemize}
	\item OneBM is the static propositionalization system on which LazyBum is based.  As the original OneBM could not be made available to us, we implemented our own version, which shares the same code base as LazyBum. As a result, our OneBM and LazyBum versions are able to generate the same features.
	\item MODL~\cite{Boulle2014,Boulle2018} is a recent static propositionalization approach included in the Khiops data mining tool.
	
	\item nFOIL~\cite{Landwehr2005}. Like LazyBum, nFOIL performs dynamic propositionalization. However, nFOIL uses a naive Bayes learner instead of a decision tree to guide its search for features. nFOIL uses conjunctive clauses to represent features, while LazyBum uses simple aggregation functions.
	
	\item Wordification~\cite{PEROVSEK20156442} is another recent static propositionalization method. Each instance in a dataset is viewed as a text document, with as words the constructed features. Wordification converts each instance in a feature vector using a bag-of-words representation for its corresponding document.
	\item Tilde~\cite{Blockeel:1998:TIF:1643275.1643308} is a relational decision learner; it produces a model but no propositionalization of the data. Since LazyBum is inspired by Tilde and uses a decision tree learner to guide its feature construction, we compare with Tilde as a baseline. 
	
\end{itemize}

For each of the systems, we performed 10-fold cross-validation. The same 10 folds were used for all systems except for nFOIL and Tilde. For nFOIL and Tilde, we used their builtin 10-fold cross-validation, which choose their own 10 folds. On each dataset, we measure both predictive accuracy and run-time performance. However, OneBM, MODL and Wordification are static propositionalization approaches. They only flatten the database into a table without building a predictive model, while LazyBum also learns a decision tree. To compare predictive accuracy for these methods with LazyBum, we learn a single decision tree on their output tables. For OneBM and MODL, we used WEKA's C4.5 decision tree implementation. For Wordification, we used a default scikit-learn tree. 

To compare run-time performance, we measure the model induction time, averaged over the different folds. For OneBM, MODL and Wordification, this includes both the propositionalization time and the time to learn a decision tree.

LazyBum and OneBM were run with their default parameter settings. For LazyBum, this corresponds $MinIG=0.001$, \(MaxDepth = \infty\) an \(MinInst = 3\) (see algorithm~\ref{alg:main-LazyBum}). LazyBum and OneBM share their feature generation code, which uses default thresholds \(\mathit{DomSize}_{abs} = 40\) and \(\mathit{DomSize}_{rel} = 0.2\) for the ``contains'' aggregation function.

For MODL, the number of constructed features was set to \(1000\). Its default feature construction rules were used, without recoding the categorical or numerical features, while keeping the initial target table attributes as features. 

For Wordification, we based our experiments on the included sample scripts, using the default weighting method.

Both nFOIL and Tilde were used with their default options. As input, nFOIL expects a list of ground facts, together with a language bias of types and refinement mode definitions. The datasets were converted by using each table row as a predicate instance. In the rmode definitions used, primary and foreign key attributes were marked as possible input variables (on which unification can be performed), the other attributes were marked as output variables. If a regular column has at most five different values, it was added as a possible selection condition to the rmodes. For Tilde, we used the same language bias as for nFOIL.

\subsection{Results}
\begin{figure}[t]
	\centering
	\subfloat[b][Accuracies.\label{fig:experiments-accuracies}]{
		\includegraphics[width=0.7\textwidth]{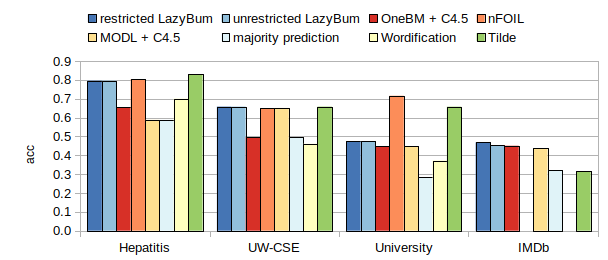}
	}
	~ 
	
	\subfloat[b][Run times relative to restricted LazyBum.\label{fig:speedups_restricted_lazybum}]{
		\includegraphics[width=0.7\textwidth]{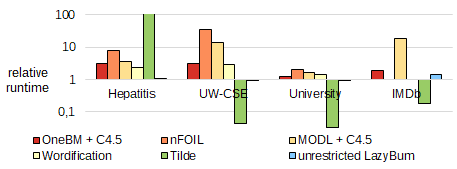}}
	\caption{Accuracy and run-time measurements for each of the datasets. For OneBM, MODL and Wordification, this includes both propositionalization and learning a tree using C4.5 (averaged over the 10 folds).}
	\label{fig:experiments-results}
\end{figure}
Before discussing our results, note that we had to modify the nFOIL setup for two datasets. For Hepatitis, nFOIL ran for four days without finishing when using a language bias containing constants. Hence, nFOIL's reported results for Hepatitis use a language bias without constants. For IMDb, the largest of the datasets, nFOIL always crashed, and Wordification did not succeed in propositionalising the first fold in eight hours, after which it was canceled. At that point, it was using $15.5$ gigabytes of memory.

\paragraph{Predictive accuracy}
Figure~\ref{fig:experiments-accuracies} shows the accuracies of the different approaches, together with the majority class frequency. Tilde has the highest accuracy on all datasets, except for IMDb. When comparing propositionalization methods, both LazyBum approaches have the highest accuracy on the UW-CSE and IMDb datasets. On Hepatitis, both LazyBum versions are almost as accurate as nFOIL, and they outperform the static approaches. 

On University, the smallest of our datasets, nFOIL and Tilde noticeably outperform all other approaches. Inspecting the nFOIL models for University shows that a large part of the generated feature clauses contain multiple instances of some predicate. That is, features contain self-joins of tables. The Tilde trees also show the same predicate being used multiple times along a branch. In comparison, our LazyBum and OneBM implementations only allow each table to occur once in a join path; they cannot generate these features. This may be why nFOIL performs better on University.

It is noteworthy that LazyBum outperforms OneBM+C4.5 on all datasets. The most likely reason is that OneBM generates so many features that it harms the performance of C4.5.

\paragraph{Run-time performance}	
Figure~\ref{fig:speedups_restricted_lazybum} shows the run time of the tested approaches relative to the restricted and unrestricted versions of LazyBum. Tilde is faster than all other approaches on all datasets but Hepatitis, for which it was $665$ times slower. This is likely due to the high number of refinements for each clause, as nFOIL did not even finish without modifying the language bias. Possibly contributing to Tilde's relative speed is that it does not propositionalize.

When comparing between the propositionalization methods, the restricted LazyBum is between $1.2$ and $35.8$ times faster than its competitors. The smallest speedups are for the University dataset, which is substantially smaller than the other datasets. Remarkably, the restricted LazyBum is only noticeably faster than the unrestricted version on the IMDb dataset. As IMDb is the largest dataset, there is the most to gain from using fewer joins. For the smaller University and UW-CSE datasets, the restricted version is slightly slower due to having a more complex extension strategy.

\paragraph{General discussion}
In summary, LazyBum always results in significant run time improvements while still achieving equivalent predictive performance on three of the four datasets. Note that for the OneBM, MODL and Wordification settings, most of the time is spent building the data table, with the decision tree induction time being almost negligible in comparison. LazyBum has the advantage of possibly not having to propositionalize the whole table. 

\section{Conclusion}
\label{sec:conclusion}

We have implemented LazyBum, a lazy version of OneBM, a recently proposed system for propositionalizing relational databases, and evaluated its performance relative to OneBM and to several other propositionalization methods.  Our experimental results suggest that LazyBum outperforms all other systems in terms of speed, sometimes by an order of magnitude, and this usually without significant loss of accuracy (the one exception being nFOIL on the Hepatitis data). Moreover, LazyBum systematically outperforms OneBM in terms of accuracy, which can only be explained by the fact that OneBM's eager generation of (many irrelevant) features is harmful to the subsequent learning process.  These results suggest that lazy propositionalization by interleaving a decision tree learner with the feature generation process is an effective approach to mining relational data.

\section{Acknowledgments}
Work supported by the KU Leuven Research Fund (C14/17/070, ``SIRV''), Research Foundation -- Flanders (project G079416N, MERCS), and the Flanders AI Impulse Program.
The authors thank Marc Boull\'e for his responsiveness and help with the Khiops system.

\bibliographystyle{splncs04}
\bibliography{main}

\end{document}